\documentclass[10pt,twocolumn,letterpaper]{article}

\usepackage[pagenumbers]{cvpr} 

\usepackage{amsmath}
\usepackage{amssymb}
\usepackage{booktabs}
\usepackage{times}  
\usepackage{helvet}  
\usepackage{courier}  
\usepackage[hyphens]{url}  
\usepackage{graphicx} 
\usepackage{caption} 
\DeclareCaptionStyle{ruled}{labelfont=normalfont,labelsep=colon,strut=off} 
\usepackage{algorithm}
\usepackage{algorithmic}

\usepackage{url}
\usepackage{amsfonts}
\usepackage{amsmath}
\usepackage{multirow} 
\usepackage{color}
\usepackage{graphicx}

\newcommand{\model}{CCF}
\newcommand{\option}[3]{\ifthenelse{\equal{#1}{old}}{#2}{\textcolor{blue}{#3}}}

\usepackage[pagebackref,breaklinks,colorlinks]{hyperref}

\usepackage[capitalize]{cleveref}
\crefname{section}{Sec.}{Secs.}
\Crefname{section}{Section}{Sections}
\Crefname{table}{Table}{Tables}
\crefname{table}{Tab.}{Tabs.}

\def\cvprPaperID{8267} 
\def\confName{CVPR}
\def\confYear{2022}

\begin{document}

\title{Exploring Category-correlated Feature for Few-shot Image Classification}

\author{Jing Xu\textsuperscript{1},~~Xinglin Pan\textsuperscript{2},~~Xu Luo\textsuperscript{2},~~Wenjie Pei\textsuperscript{1},~~Zenglin Xu\textsuperscript{1}\\
\textsuperscript{1}Harbin Institute of Technology, Shenzhen\\
\textsuperscript{2}University of Electronic Science and Technology of China\\
{\tt\small xujing.may@gmail.com}
}
\maketitle

\begin{abstract}
   Few-shot classification aims to adapt classifiers to novel classes with a few training samples. However, the insufficiency of training data  may cause a biased estimation of feature distribution in a certain class. To alleviate this problem, we present a simple yet effective feature rectification method by exploring the category correlation between novel and base classes as the prior knowledge.  We explicitly capture such correlation by mapping features into a latent vector with dimension matching the number of base classes, treating it as the logarithm probability of the feature over base classes. Based on this latent vector, the rectified feature is directly constructed by a decoder, which we expect maintaining category-related information while removing other stochastic factors, and consequently being closer to its class centroid. Furthermore, by changing the temperature value in softmax, we can re-balance the feature rectification and reconstruction for better performance. Our method is generic, flexible and agnostic to any feature extractor and classifier, readily to be embedded into existing FSL approaches. Experiments verify that our method is capable of rectifying biased features, especially when the feature is far from the class centroid. The proposed approach consistently obtains considerable performance gains on three widely used benchmarks, evaluated with different backbones and classifiers.
   The code will be made public.

\end{abstract}

\section{Introduction}

Relying on massive labeled image data, deep networks, especially Convolutional Neural Networks (CNNs)~\cite{DBLP:conf/nips/KrizhevskySH12, DBLP:journals/corr/SimonyanZ14a,DBLP:conf/cvpr/HeZRS16} have achieved great success in image classification tasks. However, the need of large-scale data for training cannot be always satisfied when the data itself is hard to obtain (e.g., medical images, satellite images, or new species of insects) or the labor needed for labeling is prohibitive to afford. Consequentially, when not provided with enough amounts of data, deep networks like CNNs suffer from severe 
deterioration of generalization ability. To address such drawbacks, 
Few-shot Learning (FSL)~\cite{DBLP:conf/icml/FinnAL17,DBLP:conf/nips/VinyalsBLKW16,DBLP:conf/nips/SnellSZ17} has been resurrected recently as a more practical and challenging task, aiming at adapting the classifier learned from a relatively large \emph{base} dataset to \emph{novel} classes of images with extremely scarce labeled data.

\begin{figure}[t!]
\centering    
\includegraphics[width=0.85\columnwidth]{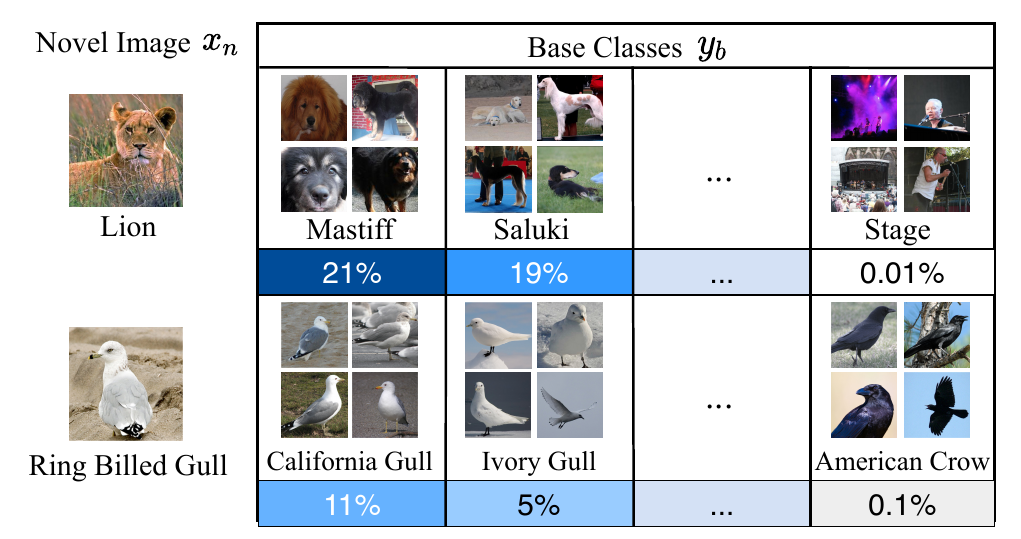}  
\caption{The motivation of the proposed method. Given a novel image $x_n$, there exists concept relationship $p(y_b|x_n)$ between the image and base classes $y_b$. 
} 
\label{fig:motivation}     
\end{figure}

\begin{figure*}[t]
  \centering
  
  \begin{subfigure}{0.62\linewidth}
    \includegraphics[width=0.95\columnwidth]{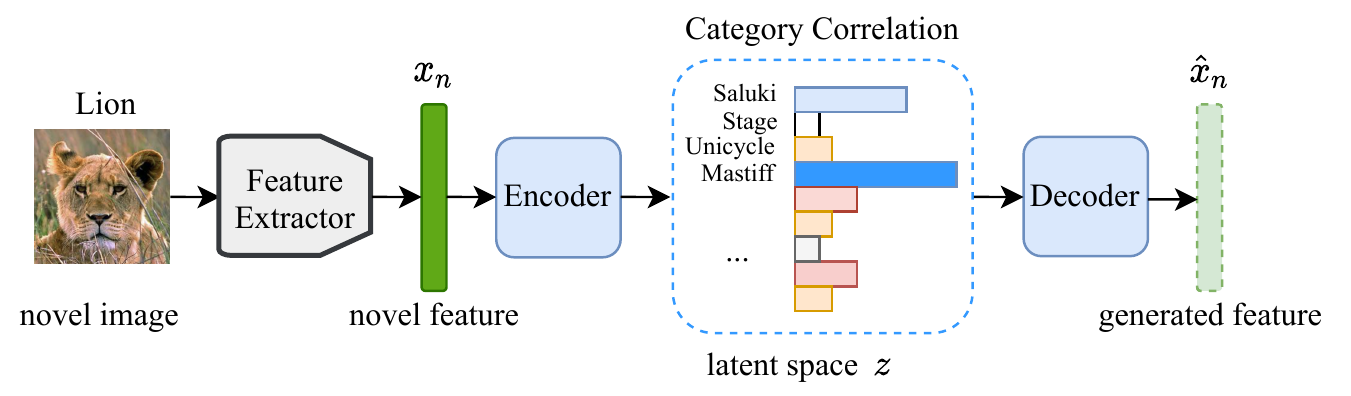}
    \caption{}
    \label{fig:short-a}
  \end{subfigure}
  \begin{subfigure}{0.36\linewidth}
   \includegraphics[width=1\columnwidth]{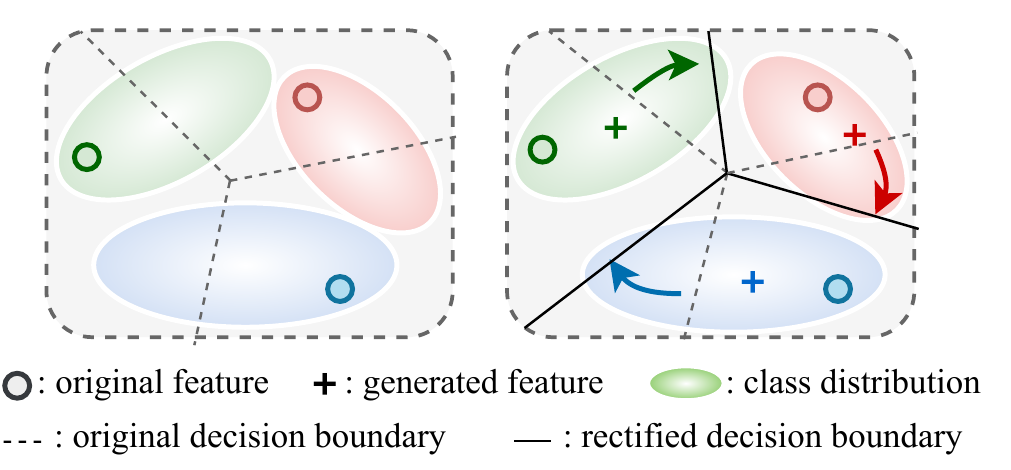}
    \caption{}
    \label{fig:short-b}
  \end{subfigure}
 
  \caption{(a) Illustration of the feature correction process. Given an image from novel class ``Lion", after using the pre-trained feature extractor, the corresponding feature is fed into the Encoder and getting the latent vector $z$ encoded category relationship among base categories. Given $z$, the rectified feature $\hat{x}$ is generated by the Decoder. (b) The decision boundary may be biased estimated by the original features when they are far away from its ground-truth centers. By combine the original and the rectified features, the decision boundary can be rectified to cover the real class distribution more accurately. }
  \label{fig:tempo_T}
  
\end{figure*}

Most studies for FSL focus on how to appropriately leverage the base dataset for better performance on novel few-shot tasks, including two predominant lines of works: optimization-based~\cite{DBLP:conf/icml/FinnAL17,DBLP:conf/iclr/RusuRSVPOH19} and metric-based~\cite{DBLP:conf/nips/SnellSZ17,DBLP:conf/nips/VinyalsBLKW16,DBLP:conf/cvpr/SungYZXTH18} methods.
Despite taking advantage of labeled data from base classes, both of the two approaches are easy-to-be-biased. Few or even a single image from one novel class is insufficient for the classifier to have a complete and comprehensive knowledge of the corresponding category, since high intra-class variations (e.g. background interference, occlusions, limited viewpoint) are likely to exist and the data points may be far away from their ground-truth class centroids, leading to a biased estimate of the classification boundary. Recently, a number of work introduce some sort of prior knowledge to rectify the biased data, e.g., leveraging class-level part information, attribute annotations or allowing the classifier to see all unlabeled query data during testing~\cite{DBLP:conf/cvpr/ZhangLYHZ21, DBLP:conf/eccv/LiuSQ20}.

 In real-world applications of Few-shot learning, prior knowledge of a new category is not always accessible (we even do not know its name). On the contrary, we have enough knowledge of base classes with their abundant labeled images. This source of knowledge is always underutilized, used only for the construction of the feature extractor and abandoned after training. A given novel category may exhibit similar patterns with one or few base classes, which could potentially help classifier construct a clearer recognition of the class; see Figure~\ref{fig:motivation} for an illustrative example. This inspires us to make use of category correlations of each novel feature as a pseudo-prior information for constructing a rectified feature that is less biased and more representative of the whole class, i.e., more close to the class centroid.



To this end, we propose Category-correlated Feature Corrector~(\model), a method that augments a given novel feature based on its category correlation with base classes, generating a rectified novel feature. To be specific, the \model{} is an autoencoder that explicitly models category correlation in the latent vector. This is done by matching the dimension of the latent vector with the number of base classes and  directly treating them as a measure of similarity of the feature with base classes. To achieve this, in addition to the common reconstruction loss used in autoencoders, a base-class classification loss is added on the latent vector, requiring that a softmax function on the latent vector gives a proper probability of the feature over all base classes. In this way, the latent vector is restricted to encode more semantic category-related information, and discard some category-irrelevant factors which not change the class probability of the feature. This essentially pushes the generated feature closer to class centroid.
In our experiments, this property of our model is not only observed in training, but also generalizes to novel category at testing. Given a novel image, as shown in Figure~\ref{fig:short-a}, the correlations with base classes can help to generate rectified feature to reduce estimation bias of the category distribution. 

Additionally, we find that it is important to balance the amount of category-irrelevant information in the latent space. A too strong classification loss results in a over-clustered feature space and the generated feature deviates too much from the original one; whereas mere reconstruction is not capable of suppressing useless intra-class information. We analyse and reformulate the cross-entropy loss, finding that the temperature parameter of the softmax function plays a key role in balancing the information. Thus we carefully select its value to reach a better trade-off. At evaluation, By combining each of the original and rectified support features together, we obtain a rectified decision boundary that is a more accurate approximation of the ground truth distribution compared with that uses the biased original ones only, as shown in Figure~\ref{fig:short-b}.

We validate our model by performing extensive experiments on multiple benchmark datasets. Our model achieves stable improvement on three widely used benchmarks and sets a new record on mini-ImageNet. More importantly, our method is generic, flexible and agnostic to any feature extractor and classifier, readily to be embedded into existing FSL approaches. This is verified by experiments of our model applied on different feature extractors and classifiers with stable $2\%-9\%$ improvement on 5-way 1-shot task.

\section{Related Works}

  

Few-shot learning task aims to recognize novel categories with limited labeled data, while given abundant training examples for the base classes. To overcome the data efficiency issue, optimization based and metric learning based methods have been proposed. Optimization based methods tackle the few shot problem by ``learning to learn''. MAML~\cite{DBLP:conf/icml/FinnAL17,DBLP:conf/nips/FinnXL18}, Reptile\cite{nichol2018reptile} and LEO\cite{DBLP:conf/iclr/RusuRSVPOH19} aim to learn good model initialization so that the model can achieve rapid adaption on novel classes with a limited number of labeled examples. Metric learning methods attempt to map the images into a high dimensional embedding space and perform classification by comparing the distance to the representatives of each novel class, e.g., MatchingNet\cite{DBLP:conf/nips/VinyalsBLKW16}, ProtoNet~\cite{DBLP:conf/nips/SnellSZ17} and Relation Network\cite{DBLP:conf/cvpr/SungYZXTH18}. Recently, Some papers~\cite{DBLP:conf/iclr/ChenLKWH19, DBLP:conf/iclr/DhillonCRS20} reveal that the standard transfer learning paradigm can achieve surprisingly competitive performance in Few-shot learning. The paper \cite{DBLP:conf/eccv/LiuCLL0LH20} introduces a negative margin loss to reduce the discriminability on base classes and novel classes. Inspired by some newly proposed self-supervised learning methods, MoCo~\cite{DBLP:conf/cvpr/He0WXG20}, SimCLR~\cite{DBLP:conf/icml/ChenK0H20} and ByoL~\cite{DBLP:conf/nips/GrillSATRBDPGAP20}, S2M2~\cite{DBLP:conf/wacv/Mangla0SKBK20}, Inv-Equ\cite{DBLP:conf/cvpr/Rizve0KS21} and ArL~\cite{DBLP:conf/cvpr/ZhangKJLT21} propose to use the self-supervised learning with some regularization technique to learn high-quality feature representation. Those works demonstrate that the significance of powerful feature representations and build new strong baselines in few shot learning tasks. Since the proposed \model{} is agnostic to the feature extractors, it is easy to combine our model with the existing methods.

Another line of algorithms is generated based methods, which directly deal with data deficiency problem in novel classes. This class of methods generate novel samples by data augmentation methods\cite{DBLP:conf/cvpr/WangGHH18}. 
\cite{DBLP:conf/iccv/HariharanG17} utilizes a generator to extract transferable intra-class deformations between same-class pairs in training classes and use those information to synthesize samples efficiently. With a common starting point, \cite{DBLP:conf/nips/SchwartzKSHMKFG18} and \cite{DBLP:conf/nips/GaoSZZC18} utilize the Auto-Encoder(AE) and Generative Adversarial Network(GAN) which makes the training process more stable. DC~\cite{DBLP:conf/iclr/YangLX21} proposes to use the statistics of the $k$-nearest base classes to calibrate the novel data distribution while we utilize correlation information among all the base categories. Besides, the feature generated by those methods are distributed randomly while our method can generate rectified features which are more closer to its class centers. Other methods ~\cite{DBLP:conf/cvpr/XianLSA18, DBLP:conf/cvpr/XianSSA19, DBLP:journals/tip/ChenFZJXS19} use high level semantic relationships between classes as extra prior information, i.e., class-level attributes~\cite{DBLP:conf/cvpr/ZhangLYHZ21}, word2vec~\cite{DBLP:conf/nips/MikolovSCCD13, DBLP:conf/cvpr/ZhangKJLT21} or unlabeled query data~\cite{DBLP:conf/eccv/LiuSQ20} to generate novel samples. However, these models are not applicable to the situation when prior knowledge is not available. Xue  \textit{et al.}~\cite{DBLP:conf/aaai/XueW20} attempt to address the problem by learning a mapping function from samples to their centroids for feature rectification. However, learning the mapping without any prior knowledge is difficult and their improvement is limited to one-shot learning task. In our method, the category information can be learned automatically without introducing additional information.

\section{Method}



In this work, we aim to generate rectified features of \emph{novel} classes conditional on the "bottleneck" latent space features to make the generated features closer to the class centers. The bottleneck latent space not only contains effective compact representation of the distribution of features, but also the corresponding class information, the category correlations between \emph{novel} classes and \emph{base} classes. To this end, we learn the latent space by training an Auto-Encoder on base classes and injecting the class semantic correlations of base categories into the space. 


 During the evaluation process, given a novel image, firstly, we use Box-Cox\cite{box1964analysis} transformation to refine the feature distribution. The Box-Cox can improve the symmetry and the normality of features and reduce the anomalies caused by the disparities of \emph{base} and \emph{novel} class domains. Compared with \cite{DBLP:conf/iclr/YangLX21} using Tukey to reduce the skewness of distributions, Box-Cox is more general and can be used when negative values exist. Secondly, the novel feature $x_n$ is fed into an encoder, and latent feature $z$ can be seen as the category correlation among base classes.
Based on the $z$, we can generate the rectified feature which is closer to its class center in the feature space. Lastly, a prediction classifier is learned based on the original feature and the generated one together. Our proposed \model{} can generate rectified high-quality features to get better decision boundaries.

\subsection{Problem Formulation}
In Few Shot learning, we are given the base categories $\mathcal C_{base}$ and novel categories $\mathcal C_{novel}$, where the sets of classes $\mathcal C_{base}$ and $\mathcal C_{novel}$ are disjoint,  $\mathcal C_{base} \cap \mathcal C_{novel} = \emptyset$. The base categories $\mathcal C_{base}$ have abundant labeled data, $\mathcal{D}_{base} =\{(x_b^i, y_b^i)\}_{i=0}^{m_{base}}$, where $y^i_b\in C_{base}$ and the novel categories $C_{novel}$ have scarce labeled data $\mathcal{D}_{novel} =\{(x^i_n, y^i_n)\}_{i=0}^{m_{novel}}$,  where $y^i_n\in C_{novel}$.

The common way to build the FSL task is called $N$-way $K$-shot which needs to classify $N$ classes sampled from the novel set with $K$ labeled data in each class. 
The few labeled data are called the support set $\mathcal S = \{(x_n^i, y_n^i)\}_{i=0}^{N\times K}$ contains $N\times K$ labeled samples totally. The performance of models is evaluated on the query set $\mathcal Q = \{(x_n^i, y_n^i)\}_{i= N\times K}^{N\times K + N\times Q}$, where the $Q$ denotes the number of images for each class in query set, which is sampled from same $N$ classes in each episode. 

Our main purpose is to predict the category of the unlabeled sample in the query set $\mathcal Q$, based on the labeled support set $\mathcal S$ and the auxiliary dataset $\mathcal{D}_{base}$. In this work, our approach is over feature level and is independent of pre-trained backbones. 

\subsection{Category-associative Feature Corrector}

We propose to use a feature generator conditional on category correlation to get rectified novel class features to obtain a better decision boundary. We train an Auto-Encoder on the base categories in the training stage and generate new novel features in the evaluation stage. 

\begin{figure}[h]
\centering    
\includegraphics[width=0.72\columnwidth]{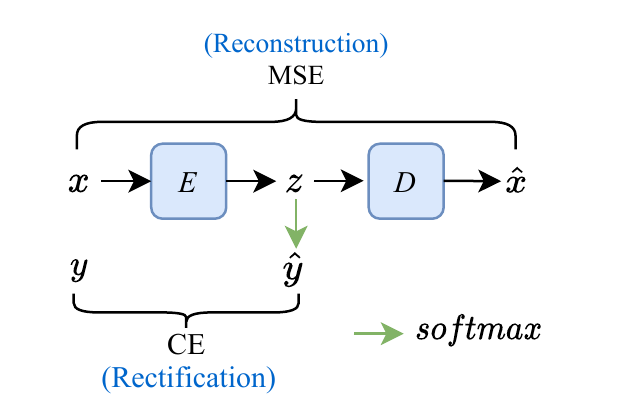}

\caption{The training process of \model{}. The MSE loss makes the generated feature $\hat{x}$ maintain the distribution of original one. By adding the CE loss between ground-truth label $y$ and predicted probability $\hat{y}$, the $z$ can be seen as log probability values over classes which incurs more category-related information in $z$.}     
\label{fig:model}     
\end{figure}

Our proposed \model{} is composed of an encoder and a decoder, while adding a cross-entropy loss for classification on the latent vector, as shown in Figure~\ref{fig:model}. The \model{} is a deep generative model which is effective in learning compact representation (the latent feature, $z$) of features with useful proprieties. Using the compact intermediate information, the mean square error(MSE) is used for feature reconstruction, which aims to generate feature, $\hat{x}$, to recovery the input, $x$, as much as possible. On the other way, the latent vector can be seen as category correlation, the logits before softmax function. The cross-entropy(CE) classification loss for rectification injects more category-level information into the latent vector, which makes the generated $\hat{x}$ closer to the class center. By finding a balance between the MSE and CE losses, the \model{} can generate high-quality rectified features which can diminish the the intra-class bias and help to learn better decision boundary.

\begin{figure*}[t]
  \centering
  
  \begin{subfigure}{0.16\linewidth}
    \includegraphics[width=1\columnwidth, trim=15 15 15 15,clip]{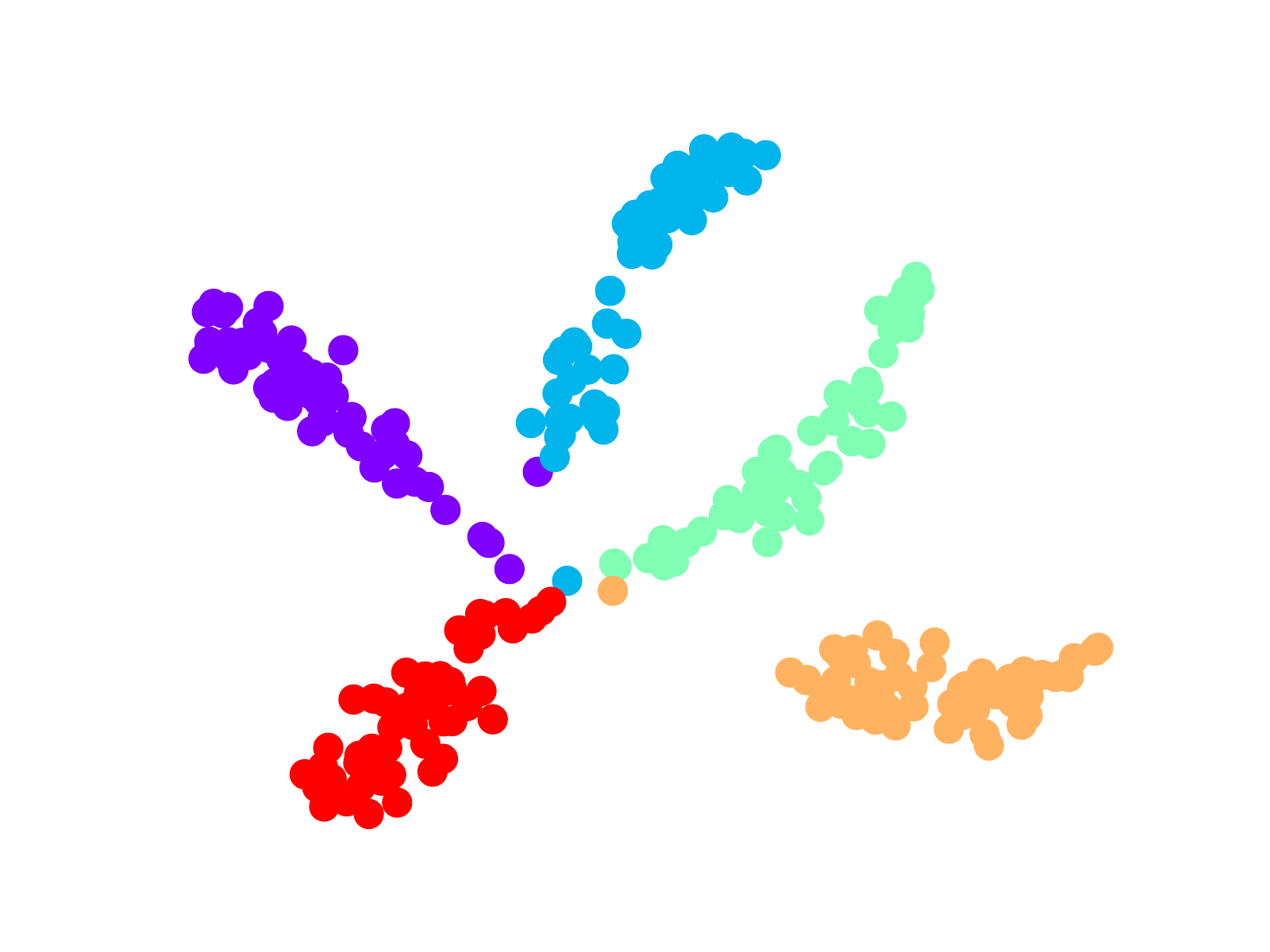}
    \caption{without CE loss}
    \label{fig:tempo_T_0}
  \end{subfigure}
  \begin{subfigure}{0.16\linewidth}
   \includegraphics[width=1\columnwidth, trim=15 15 15 15,clip]{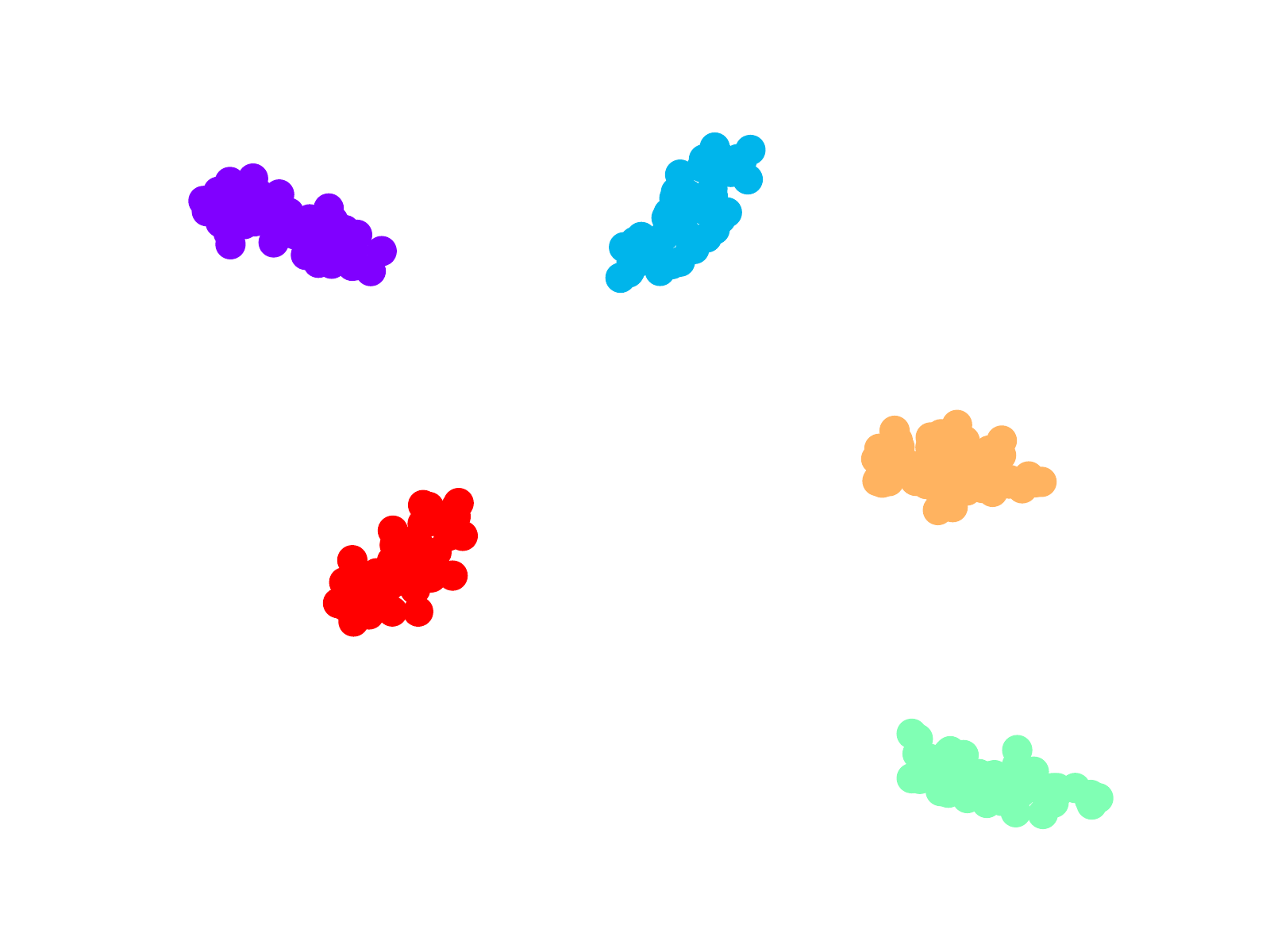}
    \caption{$T=0.05$}
    \label{fig:tempo_T_1}
  \end{subfigure}
  \begin{subfigure}{0.16\linewidth}
    \includegraphics[width=1\columnwidth, trim=15 15 15 15,clip]{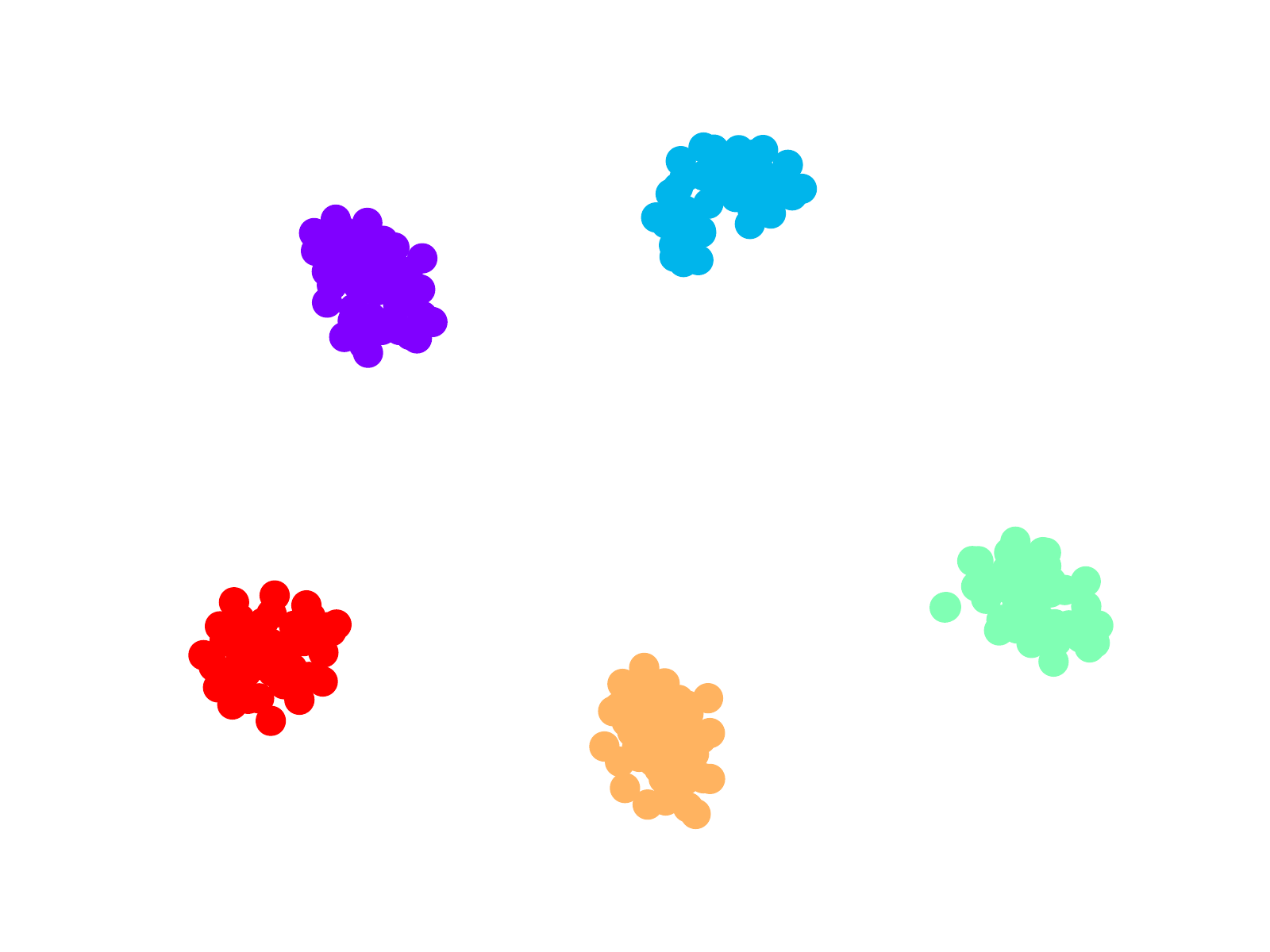}
    \caption{$T=0.1$}
    \label{fig:tempo_T_2}
  \end{subfigure}
  \begin{subfigure}{0.16\linewidth}
  \includegraphics[width=1\columnwidth, trim=15 15 15 15,clip]{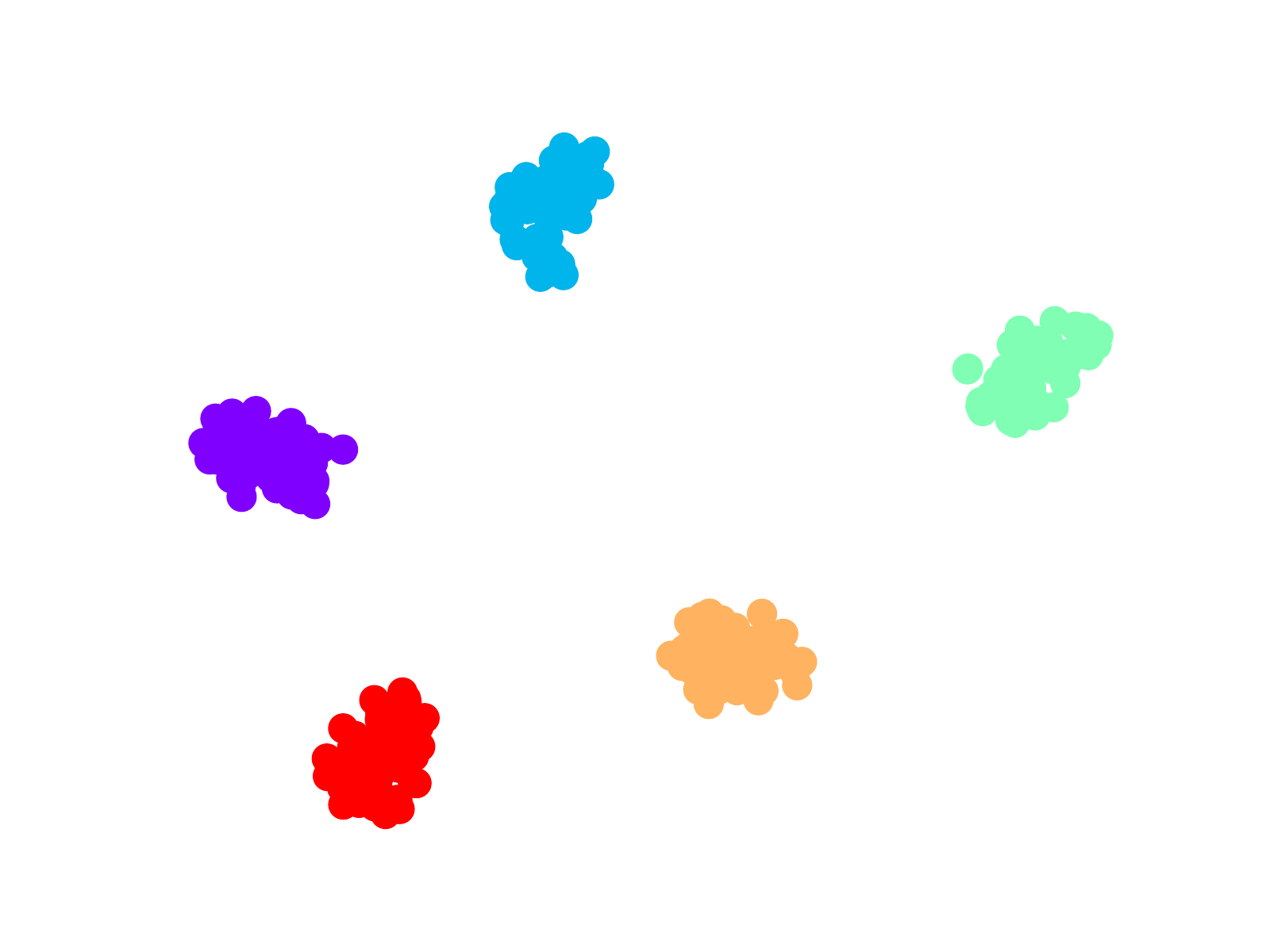}
    \caption{$T=0.5$}
    \label{fig:tempo_T_3}
  \end{subfigure}
   \begin{subfigure}{0.16\linewidth}
   \includegraphics[width=1\columnwidth, trim=15 15 15 15,clip]{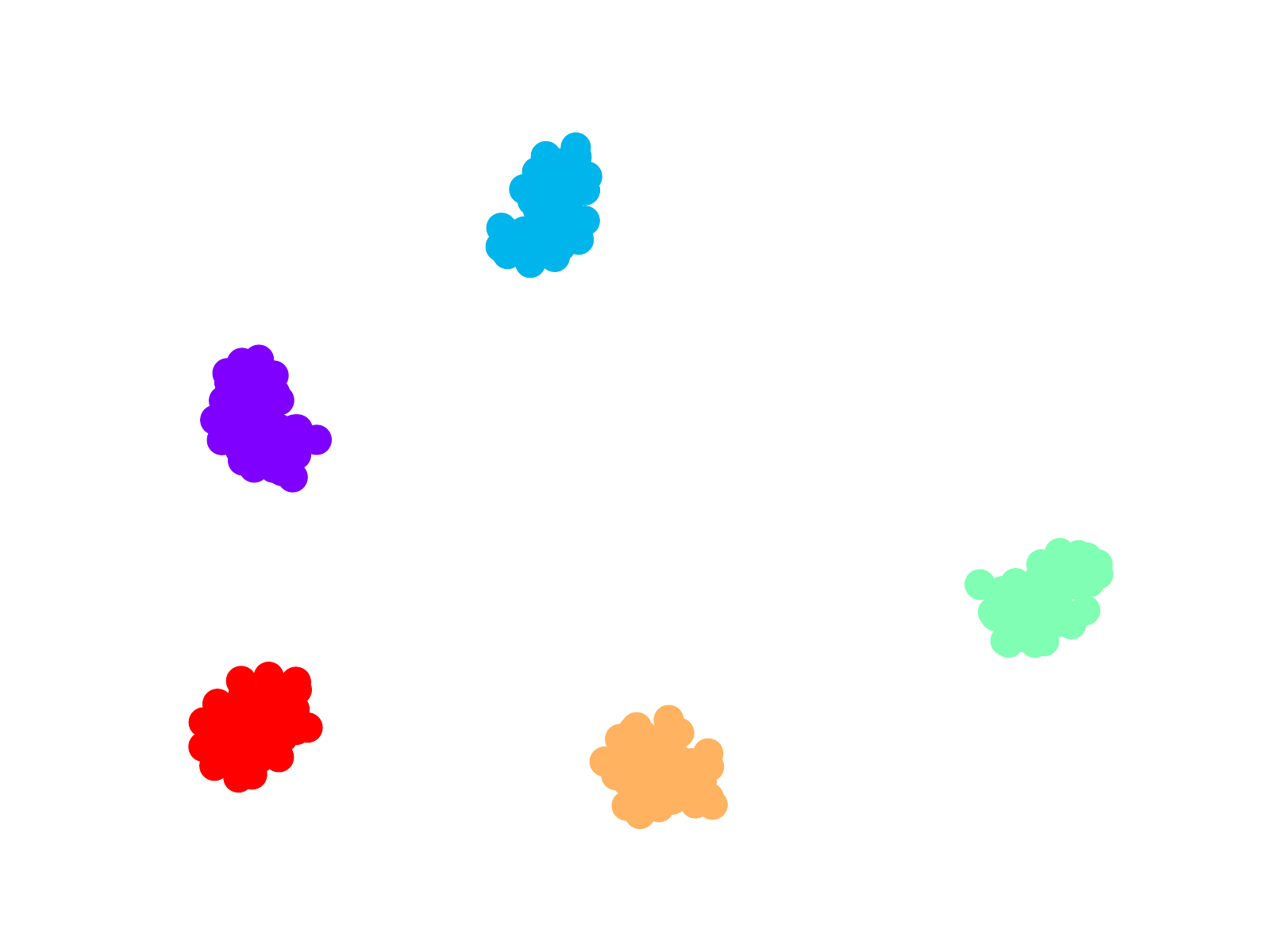}
    \caption{$T=1$}
    \label{fig:tempo_T_4}
  \end{subfigure}
   \begin{subfigure}{0.16\linewidth}
   \includegraphics[width=1\columnwidth, trim=15 15 15 15,clip]{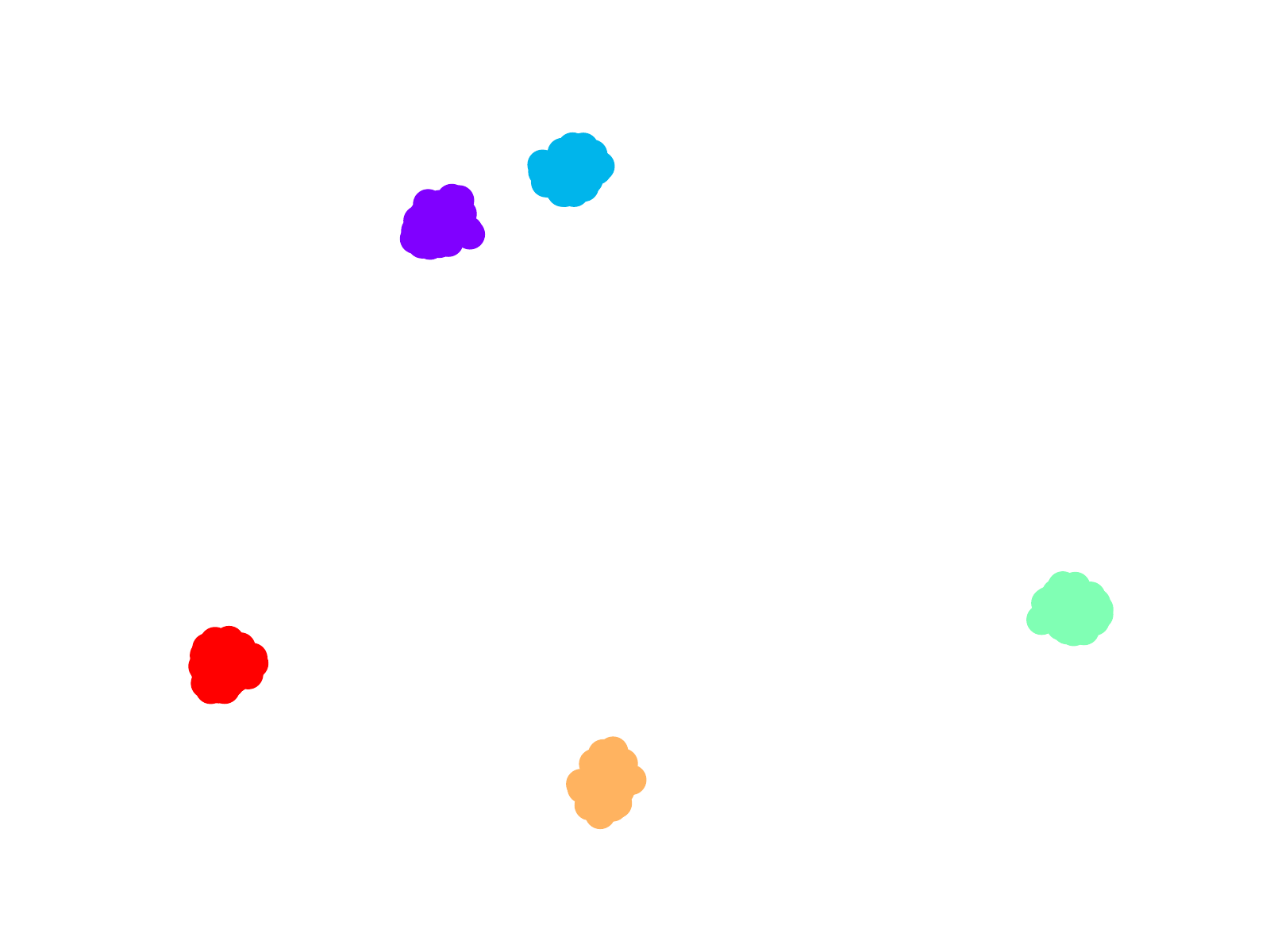}
    \caption{$T=2$}
    \label{fig:tempo_T_5}
  \end{subfigure}
  \caption{The distribution of logits vector $z$ of $5$ randomly selected classes. (a) illustrates the distribution without adding the cross-entropy classification loss. (b-f)  show the distribution with different temperature $T$. The higher $T$ makes more class information absorbed in $z$ and the features cluster tighter.}
  \label{fig:tempo_T}
  
\end{figure*}

\begin{figure}[t]
  \centering
  \includegraphics[width=0.75\columnwidth, trim=40 0 100 30,clip]{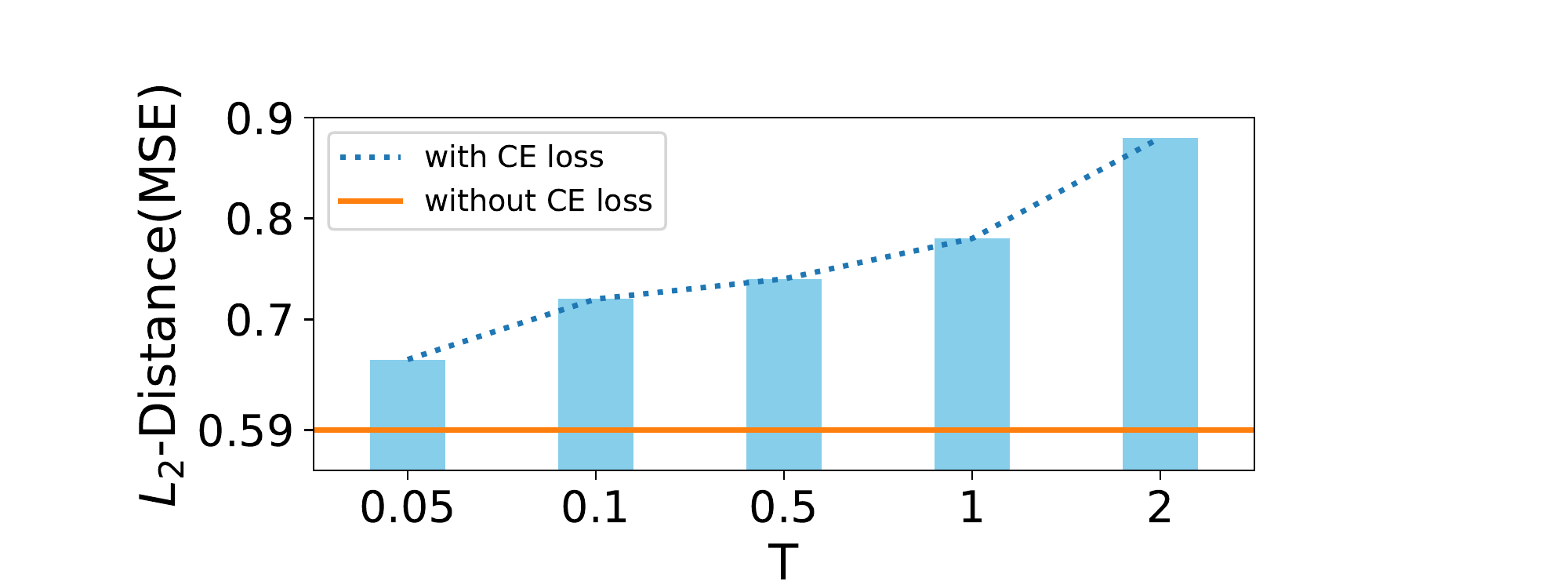}
  \caption{The $l_2$ distance (MSE loss) between original features and reconstructed ones with different temperature $T$. The higher $T$ makes the reconstructed features absorb more category information and further to the original one. }
  \label{fig:mse}
\end{figure}

\subsubsection{Training Stage}
During the training stage, our \model{} is trained on the features of base classes with sufficient labeled and capture the correlation among base classes. An encoder maps an input feature $x \in \mathcal D_{base}$ into a latent variable $z\in \mathbb{R}^{|\mathcal C_{base}|}$. The dimension of $z$ is the number of categories in base classes, $|\mathcal C_{base}|$. A decoder aims to reconstruct the feature $\hat{x}$ from the variable $z$. It has the form:

\begin{align}
z = f(x) \\ \notag
\hat{x} = g(z).
\end{align}

The function $f$ and $g$ denote the encoder and decoder, respectively. The MSE loss of \model{} for reconstruction can be formulated as:

\begin{align}
\mathcal L_{MSE} &= \sum_{x\in \mathcal D_{base}} ||x - g(f(x))||^2. 
\end{align}

 The latent space $z$ captures the underlying generating distribution of base classes features benefited from the reconstruction MSE loss. Meanwhile, the encoder $f$ also can be seen as a classifier and $z$ is the logits output (the inputs to the final softmax activation for classification). The soft label $\hat{y}$ capturing the correlation among base classes is converted by the logits, computed for each class into a probability, $\hat{y}_i = p(y_{b=i}|x)$, $i \in \{1,2, ..., |\mathcal C _{base}|\}$, by comparing $z_i$ with the other logits. Compared with probability correlation $\hat{y}$, the log probability values $z$ contains additional fine detailed relationships between classes~\cite{DBLP:conf/nips/BaC14}. The softmax with temperature $T$ can be formulated as :
 
\begin{equation}
     \hat{y}_i = \textit{softmax}(z_i/T) = \frac{e^{(z_i/T)}}{\sum_{j} e^{( z_j/T)}}.
\end{equation}

The $\hat{y}$ is supervised by the one-hot label $y$ by the CE loss:

\begin{equation}
\mathcal L_{CE} = \sum_{x\in \mathcal D_{base}} \sum_{i} y_i \text{log} (\textit{softmax}(f_i(x)/T)).
\end{equation}

The CE penalty term for feature rectification models latent space as category correlation which maintains category-related information while removes other stochastic factors. Then, the overall loss function is defined as:

\begin{equation}
\mathcal L  = \mathcal L_{MSE} + \mathcal L_{CE} + \beta ||z||^2,
\end{equation}
where the Frobenius norm penalty on $z$, $||z||^2 = \sum _i z_i^2$, can avoid over-fitting on base classes features. The hyper-parameter $\beta$ controls the strength of the regularization.


\subsubsection{Evaluation Stage}

During the evaluation stage, given an unseen feature in support set, $x_n^i \in \mathcal S$, we can get the logits feature $z$, which can be seen as the class similarity with base categories. Based on the $z$, the decoder generates new samples $\hat{x}_n^i$. Since the latent feature $\hat z$ discard some category-irrelevant information, the generated features $\hat{x}$ can be rectified closer to their class centers. we train linear classifiers by combining the original support features,  $\mathcal S = \{(x_n^i, y_n^i)\}_{i=0}^{N\times K}$, and the generated ones,  $\hat{\mathcal S} = \{(\hat{x}_n^i, y_n^i)\}_{i=0}^{N\times K}$. 


\subsection{Re-balance the Rectification and Reconstruction}
\label{sec:rebalance}
The generated features by our model not only have a similar structure to the original ones~(reconstruction) but also absorb more category-related information~(rectification). However, having the two properties simultaneously is regarded as a pair of contradictory. We can achieve better performance by re-balancing the two properties in generated features. To this end, the cross-entropy loss can be reformulated as follows:
\begin{align}
\mathcal ~ L_{CE} &= H(y, \hat{y}) \\ \notag
& = KL(y||\hat{y}) - H(y) \\  \notag 
& = KL(y||\frac{exp(z/T)}{\sum_j exp(z_j/T)})- H(y)
\end{align}
where the $H(y, \hat{y})$ and $KL(y||\hat{y})$ denote the cross-entropy and KL divergence between the distribution of the label $y$ and predicted probability label $\hat{y}$. The entropy of label $y$, denoted as $H(y)$ is a constant which do not affect the optimization process. So, minimizing the CE loss equals minimizing the KL divergence between ground truth label $y$ and latent vector $z$ which makes the latent space capture more category-related information. As shown in Figure~\ref{fig:tempo_T}, the latent vectors are more dispersed without CE loss compared with adding CE loss. The temperature T controls the smoothness of the distribution of $z$. The higher value of $T$ makes more class information absorbed in $z$.

The mean square error loss can be rewrite as:
\begin{align}
\mathcal ~ L_{MSE} = \mathbb E_x [l_2(x, g(z))]
\end{align}
where the $l_2(x, g(z))$ denotes the $l_2$ distance between $x$ and $g(z)$. Minimizing the MSE loss keeps the $z$ contains the distribution of the features in the latent space. However, injecting category-related information into $z$ will lead the reconstruction error in generated features. As shown in Figure~\ref{fig:mse}, the higher $T$ with more class information results in higher reconstruction error.


Our main purpose is to rectify original feature as much as possible while maintain the reconstruction performance. We can change the hyper-parameter $T$ to control the class information in $z$ and re-balance the degree between feature rectification and reconstruction.  In Figure~\ref{fig:tempo_T} and \ref{fig:mse}, we show how the two information changes when using different T. Using a higher value for $T$ makes the latent vector contain more class information and the model concentrate on the feature rectification. By decreasing the value, more attention has been turned to reconstruction.

\section{Experiment}

\begin{table*}[h]
\centering
\caption{Average 5-way 1-shot and 5-shot classification accuracy on miniImageNet and CUB datasets. The optimization-based methods are marked as type O, metric-based as M and generated-based as G. The top results are marked in bold. The S2M2* denotes the model re-implemented by ourselves.}
\begin{tabular}{c|c|c|c|c|c}
\hline
\multirow{2}{*}{Methods} & \multirow{2}{*}{Type} & \multicolumn{2}{c|}{miniImageNet} & \multicolumn{2}{c}{CUB} \\ \cline{3-6} 
                  &                   &    5-way 1-shot       &    5-way 5-shot      &     5-way 1-shot         &     5-way 5-shot     \\ \hline \hline
  MAML~\cite{DBLP:conf/icml/FinnAL17}   & O &  54.69 ± 0.89  &   66.62 ± 0.83  &  51.67 ± 1.81   &   70.30 ± 0.08   \\
   LEO~\cite{DBLP:conf/iclr/RusuRSVPOH19} & O &61.76 ± 0.08&77.59 ± 0.12& -- & -- \\
  \hline
  Matching Net~\cite{DBLP:conf/nips/VinyalsBLKW16}  &  M   &  63.08 ± 0.80   &  75.99 ± 0.60   &   --  & --  \\
  Prototypical Net~\cite{DBLP:conf/nips/SnellSZ17} & M & 60.37 ±  0.83 &78.02 ±  0.57 & -- & --  \\
  RelationNet~\cite{DBLP:conf/icpr/ZhuangTYMZJX18} & M &50.44 ± 0.83&70.20 ± 0.66&54.48 ± 0.93&71.32 ± 0.78 \\
  Negative-Cosine~\cite{DBLP:conf/eccv/LiuCLL0LH20} & M & 62.33 ± 0.82 & 80.94 ± 0.59 &72.66 $\pm$ 0.85 & 89.40 $\pm$ 89.40\\
  CAN~\cite{DBLP:conf/nips/HouCMSC19} & M & 64.42 $\pm$ 0.84 & 79.03 $\pm$ 0.58 & --  & -- \\
  DeepEMD~\cite{DBLP:conf/cvpr/ZhangCLS20} & M & 65.91 $\pm$ 0.82 &82.41 $\pm$ 0.56 & 75.65 $\pm$ 0.83 & 88.69 $\pm$  0.50\\
  CSEI~\cite{DBLP:conf/aaai/LiWH21a} & M & 67.59 $\pm$ 0.83 & 81.93 $\pm$ 0.36 & -- & -- \\
  ArL~\cite{DBLP:conf/cvpr/ZhangKJLT21}& M & 65.21 $\pm$ 0.58 & 80.41 $\pm$ 0.49 & 50.62 & 65.87 \\
  Inv-Equ\cite{DBLP:conf/cvpr/Rizve0KS21}& M & 67.28 $\pm$ 0.80 & 84.75 $\pm$ 0.52 & -- & -- \\ \hline
  MetaGAN~\cite{DBLP:conf/nips/ZhangCGBS18} & G & 52.71 $\pm$ 0.64 & 68.63 $\pm$ 0.67& --& -- \\
  Delta-Encoder~\cite{DBLP:conf/nips/SchwartzKSHMKFG18} & G & 59.9 &  69.7 & 69.8& 82.6\\
  TriNet~\cite{DBLP:journals/tip/ChenFZJXS19} & G &58.12 $\pm$ 1.37 & 76.92 $\pm$ 0.69 & 69.61 $\pm$ 0.46& 84.10 $\pm$ 0.35 \\
  DC~\cite{DBLP:conf/iclr/YangLX21} & G & 68.57 $\pm$ 0.55 & 82.88 $\pm$ 0.42 & 79.56 $\pm$ 0.87 & 90.67 $\pm$ 0.35\\
   \hline \hline
  Softmax classifier~\cite{DBLP:conf/iclr/ChenLKWH19} & M & 51.75 $\pm$ 0.80& 74.27 $\pm$ 0.63& 65.51 $\pm$ 0.87& 82.85 $\pm$ 0.55 \\
  Softmax classifier* & M & 51.19 $\pm$ 0.43& 73.77 $\pm$ 0.28 & 65.30 $\pm$ 0.39 & 82.63 $\pm$ 0.30 \\
  Softmax classifier* + Ours & G & \textbf{57.25 $\pm$ 0.43}& \textbf{75.50 $\pm$ 0.29}& \textbf{67.52 $\pm$ 0.41} & \textbf{83.55 $\pm$ 0.31} \\
  \hline \hline
  Cosine classifier~\cite{DBLP:conf/iclr/ChenLKWH19} & M & 51.87 ± 0.77 &75.68 ± 0.63 & 67.02 $\pm$ 0.90 & 83.58 $\pm$ 0.54\\
  Cosine classifier* & M & 52.26 ± 0.43 &75.20 ± 0.29 & 67.31 $\pm$ 0.40 & 82.98 $\pm$ 0.30\\
  Cosine classifier* + Ours & M & \textbf{56.62 ± 0.43} & \textbf{76.77 ± 0.30} & \textbf{69.02 $\pm$ 0.41} & \textbf{83.79 $\pm$ 0.32}\\
  \hline \hline
  S2M2~\cite{DBLP:conf/wacv/Mangla0SKBK20} & M & 64.93 $\pm$ 0.18 & 83.18 $\pm$ 0.11 & 80.68 $\pm$ 0.81 &  90.85 $\pm$ 0.44\\
   S2M2* & M & 65.26 $\pm$ 0.43 & 83.07 $\pm$ 0.29 & 80.70 $\pm$ 0.40 &  90.70 $\pm$ 0.30\\
   S2M2* + Ours & G & \textbf{68.88 $\pm$ 0.43} & \textbf{84.59 $\pm$ 0.30}  & \textbf{81.85 $\pm$ 0.42} & \textbf{91.58 $\pm$ 0.32} \\ \hline

\end{tabular}
\label{tab:mini_results}
\end{table*}

\begin{table}[h]
\centering
\caption{The 5-way 1-shot and 5-shot classification accuracy on miniImageNet dataset with randomly crops 9 patches for test. The best results are marked in bold.}
\begin{tabular}{c|c|c}
\hline
\multirow{2}{*}{Methods}  & \multicolumn{2}{c}{miniImageNet}  \\ \cline{2-3} 
                          &    5-way 1-shot       &    5-way 5-shot         \\ \hline \hline
  
   DeepEMD+9s~\cite{DBLP:conf/cvpr/ZhangCLS20}& 68.77 $\pm$ 0.29& 84.13 $\pm$ 0.53 \\
   CSEI+9s~\cite{DBLP:conf/aaai/LiWH21a} & 68.94 $\pm$ 0.28 & 85.07 $\pm$ 0.50 \\
   S2M2$^*$+9s &  69.05 $\pm$ 0.44 &  85.10 $\pm$ 0.28  \\
 S2M2$^*$ + Ours + 9s & \textbf{70.50 $\pm$ 0.44} & \textbf{86.28 $\pm$ 0.29} \\ \hline
\end{tabular}
\label{tab:9s_results}
\end{table}

\subsection{Implementation Details and Datasets}

\textbf{Implementation Details. }
The encoder of the \model{} is a one-layer fully-connected network of hidden dimension $2048$, varied latent dimension of $z$ depending on the number of base categories and input dimension depending on the dimension of the pre-trained feature extractor. The decoder is a single-layer fully-connected network, which takes input as latent feature $z$, and outputs reconstructed/generated features. The nonlinear activation function LeakyReLU is consistently used in both encoder and decoder. The whole structure is trained using Adam optimizer with a learning rate of $0.0001$ with early stopping based on the Few-Shot classification accuracy on the validation set. The $\beta$ of Frobenius norm on $z$ is set to be $0.05$ on mini/tiered-ImageNet and $0.1$ on CUB. We set the temperature $T=0.1$ for 1-shot task and $T=0.02$ for 5-shot task. For each evaluation task, the trained \model{} generates one new rectified feature for each novel support feature, followed by training a simple classifier on all support and generated features to learn a reliable decision boundary of different classes for predictions of query samples. We report mean accuracy as well as the $95\%$ confidence interval on $2000$ randomly generated episodes.

\textbf{Datasets. }
We evaluate our method on three widely used datasets in FSL: \emph{mini}ImageNet\cite{DBLP:conf/iclr/RaviL17}, \emph{tiered}ImageNet\cite{DBLP:conf/iclr/RenTRSSTLZ18} and CUB\cite{WelinderEtal2010}. 

\textbf{\emph{mini}ImageNet}
is a subset of ILSVRC-12~\cite{DBLP:journals/ijcv/RussakovskyDSKS15}, including 600 images from each of 100 distinct classes of animals or objects. The categories are split into 64, 16, 20 classes for training, validation
and evaluation respectively, following previous work \cite{DBLP:conf/iclr/RaviL17}.

\textbf{\emph{tiered}ImageNet}
is also a subset derived from ILSVRC-12 but is much larger and is more challenging. It is made up of 779,165 images from 608 classes sampled from a hierarchical category structure. We adopt 351 classes as base categories, 97 as validation categories and 160 as novel categories as suggested in \cite{DBLP:conf/iclr/RenTRSSTLZ18}.

\textbf{CUB}
is a fine-grained dataset consisting of 11,788 images from 200 bird classes. Following \cite{DBLP:conf/iclr/ChenLKWH19}, we spilt the dataset into 100, 50 and 50 categories as base, validation and test categories.

\subsection{Comparisons with State-of-the-Art Methods}

\begin{table}[t]
\caption{5-way (1/5)-shot classification accuracy on tiered-ImageNet datasets. The best results are marked in bold. The S2M2* denotes the model re-implemented by ourselves.}
\begin{tabular}{c|c|c|c}
\hline
 Datset& Type & 5-way 1-shot & 5-way 5-shot \\ \hline \hline
 MAML~\cite{DBLP:conf/icml/FinnAL17} & O & 51.67 $\pm$ 1.81 & 70.93 $\pm$ 0.08  \\ 
 LEO~\cite{DBLP:conf/iclr/RusuRSVPOH19} & O & 66.33 $\pm$ 0.05&  81.44 $\pm$ 0.09\\ 
 MetaOpt~\cite{DBLP:conf/cvpr/LeeMRS19} & O& 65.81 $\pm$  0.74 & 81.75 $\pm$ 0.53\\
 CSEI~\cite{DBLP:conf/aaai/LiWH21a}  & M & 72.57 $\pm$ 0.95 & 85.72 $\pm$ 0.63 \\ 
 DeepEMD~\cite{DBLP:conf/cvpr/ZhangCLS20} & M & 71.00 $\pm$ 0.32 & 85.01 $\pm$ 0.67 \\ 
 CAN~\cite{DBLP:conf/nips/HouCMSC19} & M & 70.65  $\pm$ 0.99 & 84.08 $\pm$ 0.68 \\
 Inv-Equ\cite{DBLP:conf/cvpr/Rizve0KS21} & M & 72.21  $\pm$ 0.90 & 87.08 $\pm$ 0.58 \\
  
 S2M2* & M & 73.02 $\pm$ 0.49 & 86.98 $\pm$ 0.25 \\
 S2M2* + Ours & G & \textbf{75.01 $\pm$ 0.50} &  \textbf{87.65 $\pm$ 0.26}\\ 
 \hline
\end{tabular}
\label{tab:tiered_results}
\end{table}

In this subsection, we report the results of our proposed methodology compared with other State-of-the-Art methods.  We split them into 3 groups, optimization-based(O), metric-based(M), and generated-based(G) methods as introduced in the Related Works section.

The average 5-way 1/5-shot classification accuracy with 95\% confidence intervals on miniImageNet and CUB datasets are shown in Table~\ref{tab:mini_results}. The miniImageNet and CUB datasets have distinct category characteristics. miniImageNet consists of a wide range of classes ranging from animals to objects with a considerably low category similarity, while in CUB there exists close correlations among fine-grained bird classes. Our proposed method can improve the performance on both miniImageNet and CUB, showing the ability to generate high-quality features and capture the category correlation with different similarity granularity. We combine our \model{} with three algorithms, Softmax classifier~\cite{DBLP:conf/iclr/ChenLKWH19}, Cosine classifier~\cite{DBLP:conf/iclr/ChenLKWH19} and S2M2~\cite{DBLP:conf/wacv/Mangla0SKBK20}. For a fair comparison, we re-implement the results of 
those algorithms. Compared with the baseline model S2M2, our model gives a boost to the 5-way 1/5-shot accuracy on miniImageNet by 3.63\% /1.52\%, and on CUB by 1.10\%/0.88\%, setting new records on the miniImageNet and CUB datasets. We observe similar trends on tiered-ImageNet dataset shown in Table~\ref{tab:tiered_results}. 

Meanwhile, we evaluate our proposed method combined with the new sampling strategy(9s) proposed in DeepEMD~\cite{DBLP:conf/cvpr/ZhangCLS20}. Given original images, the strategy randomly samples 9 patches with different sizes and shapes followed by resizing these patches to 84 × 84. The results are shown in Table~\ref{tab:9s_results}. We achieve remarkable 70.50\% and 86.28\% accuracies for 1/5-shot tasks on \emph{mini}ImageNet, which outperforms all the strong competitors.

\subsection{Analysis}
To better understand the properties of the generated features by our method, we analyze the distance from the class centers $x_c$, to original features $x$, and the generated ones $\hat{x}$. Meanwhile, we use t-SNE to visualize the distribution of the generated samples. The experiments are conducted on \emph{mini}Imagenet. 

\textbf{Distance with the class centers.}
We denotes the distance between the class centers $x_c$ with corresponding the original features $x$  as $d$  and with generated features as $\hat{d}$, which $ d= \Vert x -x_c \Vert_2$ and $ \hat{d}= \Vert \hat{x} -x_c  \Vert_2$. We compare the two distances in Table~\ref{tab:distance}. In both base and novel classes, the distance $\hat d$ has lower value than $d$, which means the generated features are closer to the class centers. In Figure~\ref{fig:outlier}, we visualize the distribution of original and generated features of three randomly selected classes on base and novel classes, respectively.

\begin{figure}[t]
\centering    
\includegraphics[width=0.8\columnwidth]{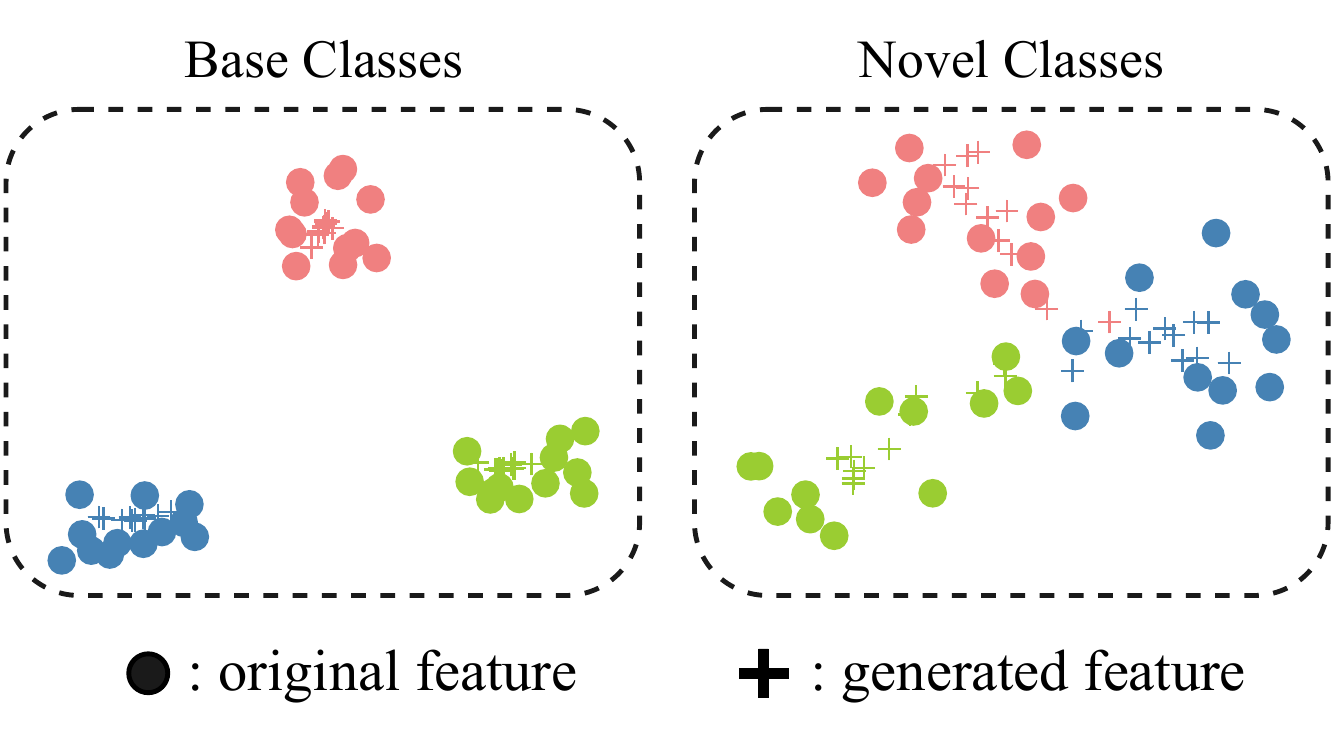}  
\caption{Visualization of the distribution of 3 randomly selected classes on base and novel classes, respectively. Compared with the original features(denoted as ``$\bullet$''), the generated features (denoted as ``+'') are closer to the class centers.} 
\label{fig:outlier}     
\end{figure}

\begin{table}[h]
\caption{Distance with the class centers. We average all the features in both base and novel classes.}
\label{tab:distance}
\centering
\begin{tabular}{c|c|c}
\hline

 $l_2$-norm distance & base & novel  \\ \hline \hline
 $d$   & 1.24 &  1.16\\  \hline
 $\hat d$ & 0.94 & 0.84 \\ \hline
\end{tabular}
\end{table}



We attribute such advantage to the class-relevant information in latent vector $z$ captured by our \model{}. We generate new features based on the latent vector which contains class semantic correlation information and somewhat discard other information in $z$, thereby removing some of the stochastic factors in data .


\textbf{Rectification of the outliers.} Our model works especially well due to the category gap existing between base and novel sets. It is likely that large intra-class variance exists in the novel feature space. This leads to the possibility of some novel support features being far away from its ground-truth center. We visually demonstrate in figure~\ref{fig:outlier} that this problem can be largely alleviated by the rectified features from our \model{}.

\begin{figure}[h]
\centering    
\includegraphics[width=0.8\columnwidth]{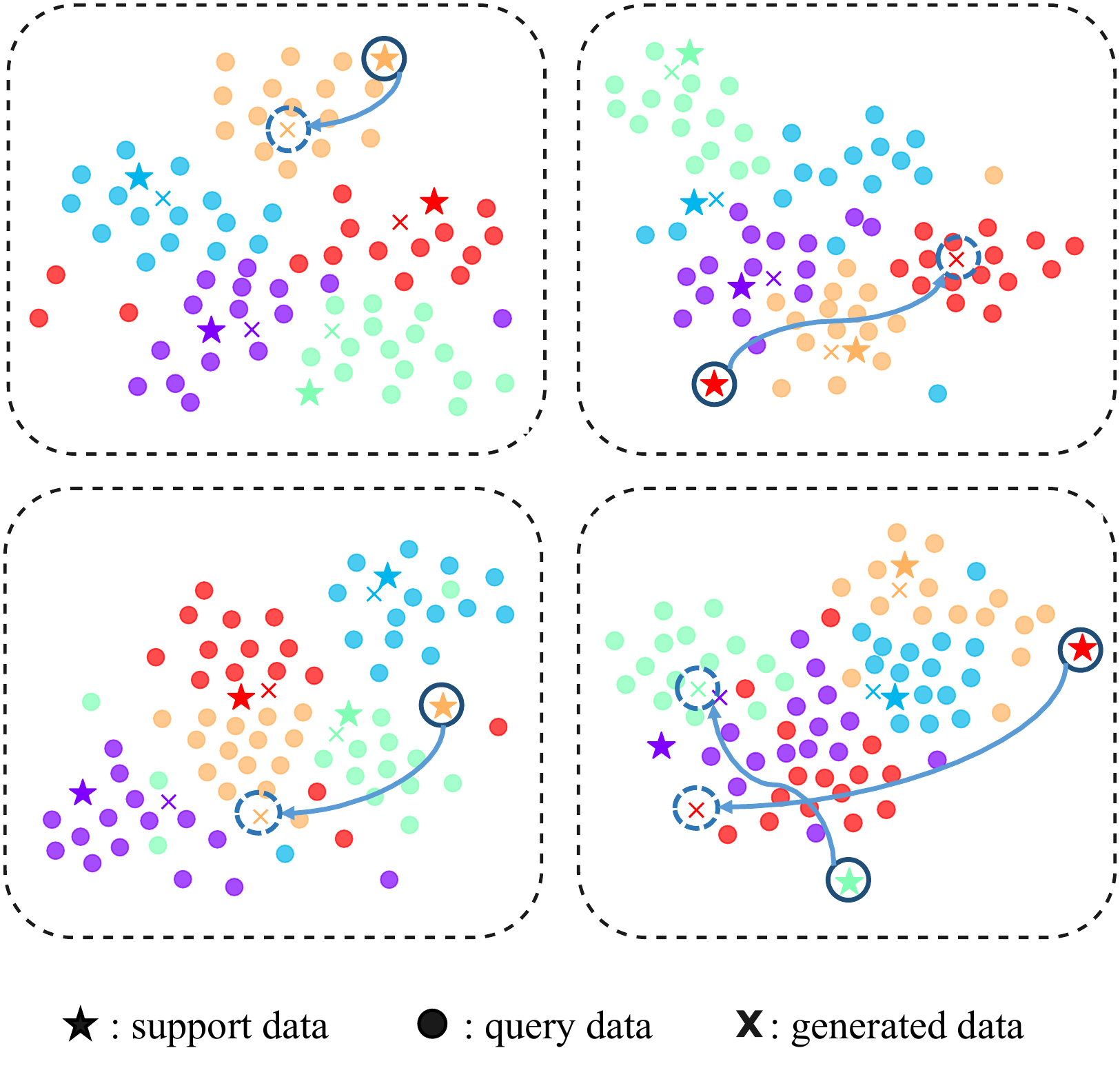}  
\caption{Visualization of the distribution of some randomly generated episodes. When the support features (denoted as ``$\star$") are far away from their ground-truth class centers, the sampled features (denoted as ``$\times$") can rectify the outliers closer to their class centers. The query feature are denoted as ``$\bullet$".} 
\label{fig:outlier}     
\end{figure}

\subsection{Ablation Study}

In this subsection, we deeply investigate the proposed method: the effect of different components and the generalization ability of our method. All experiments are conducted on mini-ImageNet.

\subsubsection{Effect of different components}

\begin{table}[h]
\caption{Ablation study on mini-ImageNet. The 5-way 1/5-shot classification accuracy(\%) are reported.}
\label{tab:ablation}
\centering
\begin{tabular}{cccc|c|c}
\hline

 B & +B-C & +AE & +\model{} & 1-shot & 5-shot  \\ \hline \hline
 $\checkmark$   & && & 65.25$\pm$0.44 & 82.85$\pm$0.29 \\ \hline

 $\checkmark$&$\checkmark$  &  & & 65.97$\pm$0.43&  83.95$\pm$0.29\\ \hline
   $\checkmark$ &$\checkmark$ &  $\checkmark$ & &   67.15$\pm$0.43 & 83.89$\pm$0.28\\   \hline
  $\checkmark$ &$\checkmark$ &&$\checkmark$ &  \textbf{68.88$\pm$0.43} & \textbf{84.59$\pm$0.30} \\ \hline
    
\end{tabular}
\end{table}


We conduct an ablation study to assess the effects of the proposed components. We use S2M2 as the feature extractor and adopt a logistic regression classifier. The results are shown in Table~\ref{tab:ablation}. The first row denotes the baseline(B) that does not utilize any augmented samples for training the classifier. After using Box-Cox(+B-C)~\cite{box1964analysis} to calibrate the distribution of features, we can get higher accuracy. The third line denotes that we only use the generated features to learn the classification boundary. Besides, we use the original Auto-Encoder(+AE) to generate new feature. Since the AE aims to learn a compact representation of feature and ignore insignificant data (“noise”), the accuracy is improved to some extent, but the improvement is limited.
After combining the original support with the generated features(+\model{}), the performance is further improved.

\subsubsection{Generalization of our method}

\begin{figure}[h]
\centering    
\includegraphics[width=1.\columnwidth, trim=30 0 20 20,clip]{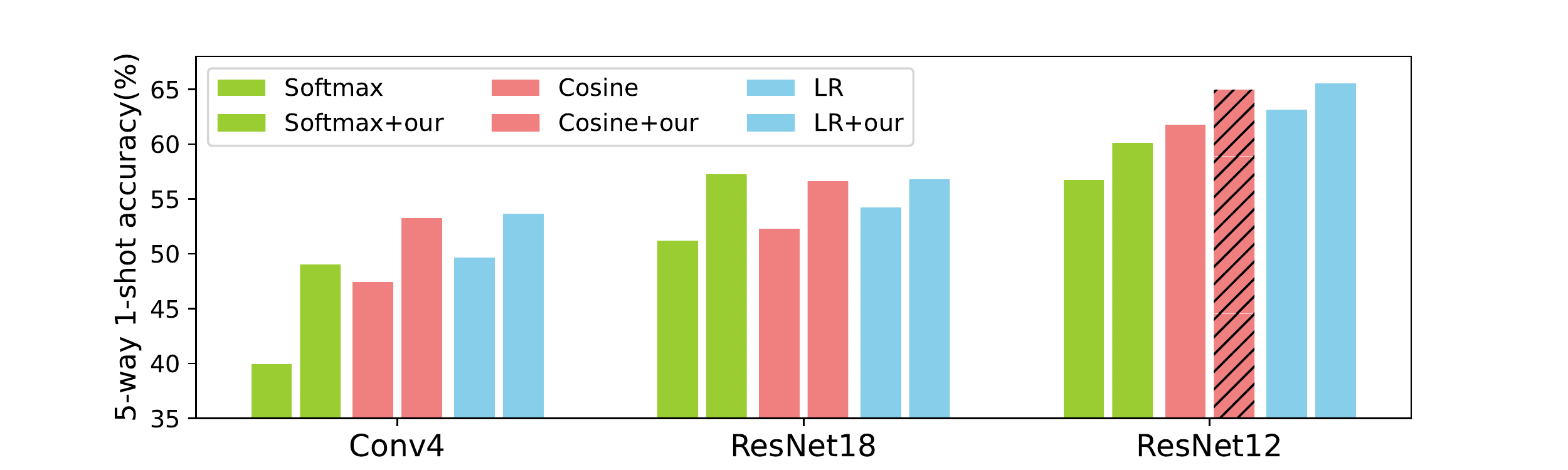}  
\caption{The histogram of 5-way 1-shot accuracies on mini-Imagenet with different backbones and classifiers.} 
\label{fig:general}     
\end{figure}

To show our method is agnostic to different feature extractors and the classifiers, we apply our method on additional three different backbones with two few-shot classification algorithms, ``Baseline'' and ``Baseline++'' proposed in~\cite{DBLP:conf/iclr/ChenLKWH19}. Besides, we combine three widely used classifiers: softmax, cosine and the logistic regression with our method to demonstrate the generalization ability of our method. We use the algorithm ``Baseline'' with softmax classifier and ``Baseline++'' with cosine and logistic regression classifiers. The results are shown in Figure~\ref{fig:general}. We get stable $2\%-9\%$ on all the backbones and classifiers.

\subsubsection{Hyper-parameters selection}

\begin{figure}[h]
\centering    
\includegraphics[width=.85\columnwidth, trim=30 0 20 20,clip]{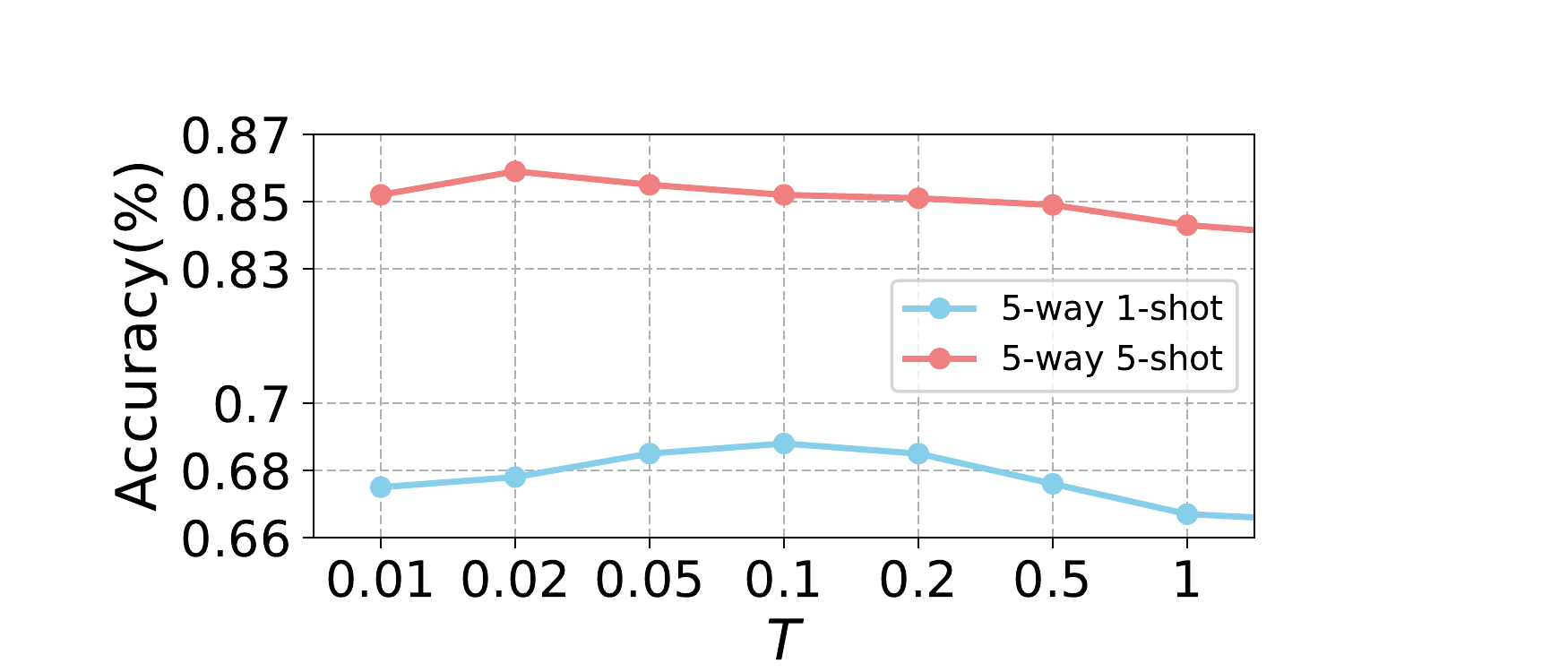}  
\caption{The accuracies of 1/5-shot task on mini-Imagenet with different temperature  $T$.} 
\label{fig:tempo_2}     
\end{figure}

As introduced in Section~\ref{sec:rebalance}, the temperature $T$ in softmax activation is important in balancing the feature rectification and reconstruction. Here we analyze how the temperate value effect the classification performance. Figure~\ref{fig:tempo_2} shows the results on mini-ImageNet on both 1/5-shot tasks. The best performance is achieved in 1-shot task when $T=0.1$ and $T=0.02$ in 5-shot task. The experiment results show that we need pay more attention for feature rectification in 1-shot task, since only one sample can lead a biased decision boundary easily. For 5-shot task, given 5 samples per class, the data-biased problem is somewhat alleviated, so, we need focus more on maintaining the original distribution of data for feature rectification.

\textbf{Limitation.} For $k$-shot learning, we need to balance the feature rectification and reconstruction and adjust the hyper-parameter $T$. With the increase of $k$, the improvement brought by our method will decrease gradually.

\section{Conclusion}

In this paper, we propose a generative model named \model{} for Few-shot classification. The \model{} is composed of an encoder and a decoder, which can capture the latent low-dimensional manifold and transferable category correlation of features on base classes and use them to generate novel image. To be specific, the \model{} is trained on base categories which encodes the feature distribution into a latent intermediate space. Meanwhile, we add a classification penalty term to encode category information into the space. Due to the penalty, the reconstructed feature outputted by the decoder is more closer to its class center, which can be seen as a rectification of the original one. After training, given a novel data, the rectified novel data is generated based on its latent encoding with the category correlation with base classes. We thoroughly verified the effectiveness of our proposed methods, which can consistently achieve the appealing performance on different datasets.

{\small
\bibliographystyle{ieee_fullname}
\bibliography{egbib}

\begin{thebibliography}{10}\itemsep=-1pt

\bibitem{DBLP:conf/nips/BaC14}
Jimmy Ba and Rich Caruana.
\newblock Do deep nets really need to be deep?
\newblock In {\em {NIPS}}, pages 2654--2662, 2014.

\bibitem{box1964analysis}
George~EP Box and David~R Cox.
\newblock An analysis of transformations.
\newblock {\em Journal of the Royal Statistical Society: Series B
  (Methodological)}, 26(2):211--243, 1964.

\bibitem{DBLP:conf/icml/ChenK0H20}
Ting Chen, Simon Kornblith, Mohammad Norouzi, and Geoffrey~E. Hinton.
\newblock A simple framework for contrastive learning of visual
  representations.
\newblock In {\em {ICML}}, volume 119 of {\em Proceedings of Machine Learning
  Research}, pages 1597--1607. {PMLR}, 2020.

\bibitem{DBLP:conf/iclr/ChenLKWH19}
Wei{-}Yu Chen, Yen{-}Cheng Liu, Zsolt Kira, Yu{-}Chiang~Frank Wang, and
  Jia{-}Bin Huang.
\newblock A closer look at few-shot classification.
\newblock In {\em {ICLR} (Poster)}. OpenReview.net, 2019.

\bibitem{DBLP:journals/tip/ChenFZJXS19}
Zitian Chen, Yanwei Fu, Yinda Zhang, Yu{-}Gang Jiang, Xiangyang Xue, and Leonid
  Sigal.
\newblock Multi-level semantic feature augmentation for one-shot learning.
\newblock {\em {IEEE} Trans. Image Process.}, 28(9):4594--4605, 2019.

\bibitem{DBLP:conf/iclr/DhillonCRS20}
Guneet~Singh Dhillon, Pratik Chaudhari, Avinash Ravichandran, and Stefano
  Soatto.
\newblock A baseline for few-shot image classification.
\newblock In {\em {ICLR}}. OpenReview.net, 2020.

\bibitem{DBLP:conf/icml/FinnAL17}
Chelsea Finn, Pieter Abbeel, and Sergey Levine.
\newblock Model-agnostic meta-learning for fast adaptation of deep networks.
\newblock In {\em {ICML}}, volume~70 of {\em Proceedings of Machine Learning
  Research}, pages 1126--1135. {PMLR}, 2017.

\bibitem{DBLP:conf/nips/FinnXL18}
Chelsea Finn, Kelvin Xu, and Sergey Levine.
\newblock Probabilistic model-agnostic meta-learning.
\newblock In {\em NeurIPS}, pages 9537--9548, 2018.

\bibitem{DBLP:conf/nips/GaoSZZC18}
Hang Gao, Zheng Shou, Alireza Zareian, Hanwang Zhang, and Shih{-}Fu Chang.
\newblock Low-shot learning via covariance-preserving adversarial augmentation
  networks.
\newblock In {\em NeurIPS}, pages 983--993, 2018.

\bibitem{DBLP:conf/nips/GrillSATRBDPGAP20}
Jean{-}Bastien Grill, Florian Strub, Florent Altch{\'{e}}, Corentin Tallec,
  Pierre~H. Richemond, Elena Buchatskaya, Carl Doersch, Bernardo~{\'{A}}vila
  Pires, Zhaohan Guo, Mohammad~Gheshlaghi Azar, Bilal Piot, Koray Kavukcuoglu,
  R{\'{e}}mi Munos, and Michal Valko.
\newblock Bootstrap your own latent - {A} new approach to self-supervised
  learning.
\newblock In {\em NeurIPS}, 2020.

\bibitem{DBLP:conf/iccv/HariharanG17}
Bharath Hariharan and Ross~B. Girshick.
\newblock Low-shot visual recognition by shrinking and hallucinating features.
\newblock In {\em {ICCV}}, pages 3037--3046. {IEEE} Computer Society, 2017.

\bibitem{DBLP:conf/cvpr/He0WXG20}
Kaiming He, Haoqi Fan, Yuxin Wu, Saining Xie, and Ross~B. Girshick.
\newblock Momentum contrast for unsupervised visual representation learning.
\newblock In {\em {CVPR}}, pages 9726--9735. Computer Vision Foundation /
  {IEEE}, 2020.

\bibitem{DBLP:conf/cvpr/HeZRS16}
Kaiming He, Xiangyu Zhang, Shaoqing Ren, and Jian Sun.
\newblock Deep residual learning for image recognition.
\newblock In {\em {CVPR}}, pages 770--778. {IEEE} Computer Society, 2016.

\bibitem{DBLP:conf/nips/HouCMSC19}
Ruibing Hou, Hong Chang, Bingpeng Ma, Shiguang Shan, and Xilin Chen.
\newblock Cross attention network for few-shot classification.
\newblock In {\em NeurIPS}, pages 4005--4016, 2019.

\bibitem{DBLP:conf/nips/KrizhevskySH12}
Alex Krizhevsky, Ilya Sutskever, and Geoffrey~E. Hinton.
\newblock Imagenet classification with deep convolutional neural networks.
\newblock In {\em {NIPS}}, pages 1106--1114, 2012.

\bibitem{DBLP:conf/cvpr/LeeMRS19}
Kwonjoon Lee, Subhransu Maji, Avinash Ravichandran, and Stefano Soatto.
\newblock Meta-learning with differentiable convex optimization.
\newblock In {\em {CVPR}}, pages 10657--10665. Computer Vision Foundation /
  {IEEE}, 2019.

\bibitem{DBLP:conf/aaai/LiWH21a}
Junjie Li, Zilei Wang, and Xiaoming Hu.
\newblock Learning intact features by erasing-inpainting for few-shot
  classification.
\newblock In {\em {AAAI}}, pages 8401--8409. {AAAI} Press, 2021.

\bibitem{DBLP:conf/eccv/LiuCLL0LH20}
Bin Liu, Yue Cao, Yutong Lin, Qi Li, Zheng Zhang, Mingsheng Long, and Han Hu.
\newblock Negative margin matters: Understanding margin in few-shot
  classification.
\newblock In {\em {ECCV} {(4)}}, volume 12349 of {\em Lecture Notes in Computer
  Science}, pages 438--455. Springer, 2020.

\bibitem{DBLP:conf/eccv/LiuSQ20}
Jinlu Liu, Liang Song, and Yongqiang Qin.
\newblock Prototype rectification for few-shot learning.
\newblock In {\em {ECCV} {(1)}}, volume 12346 of {\em Lecture Notes in Computer
  Science}, pages 741--756. Springer, 2020.

\bibitem{DBLP:conf/wacv/Mangla0SKBK20}
Puneet Mangla, Mayank Singh, Abhishek Sinha, Nupur Kumari, Vineeth~N.
  Balasubramanian, and Balaji Krishnamurthy.
\newblock Charting the right manifold: Manifold mixup for few-shot learning.
\newblock In {\em {WACV}}, pages 2207--2216. {IEEE}, 2020.

\bibitem{DBLP:conf/nips/MikolovSCCD13}
Tom{\'{a}}s Mikolov, Ilya Sutskever, Kai Chen, Gregory~S. Corrado, and Jeffrey
  Dean.
\newblock Distributed representations of words and phrases and their
  compositionality.
\newblock In {\em {NIPS}}, pages 3111--3119, 2013.

\bibitem{nichol2018reptile}
Alex Nichol and John Schulman.
\newblock Reptile: a scalable metalearning algorithm.
\newblock {\em arXiv preprint arXiv:1803.02999}, 2(3):4, 2018.

\bibitem{DBLP:conf/iclr/RaviL17}
Sachin Ravi and Hugo Larochelle.
\newblock Optimization as a model for few-shot learning.
\newblock In {\em {ICLR}}. OpenReview.net, 2017.

\bibitem{DBLP:conf/iclr/RenTRSSTLZ18}
Mengye Ren, Eleni Triantafillou, Sachin Ravi, Jake Snell, Kevin Swersky,
  Joshua~B. Tenenbaum, Hugo Larochelle, and Richard~S. Zemel.
\newblock Meta-learning for semi-supervised few-shot classification.
\newblock In {\em {ICLR} (Poster)}. OpenReview.net, 2018.

\bibitem{DBLP:conf/cvpr/Rizve0KS21}
Mamshad~Nayeem Rizve, Salman~H. Khan, Fahad~Shahbaz Khan, and Mubarak Shah.
\newblock Exploring complementary strengths of invariant and equivariant
  representations for few-shot learning.
\newblock In {\em {CVPR}}, pages 10836--10846. Computer Vision Foundation /
  {IEEE}, 2021.

\bibitem{DBLP:journals/ijcv/RussakovskyDSKS15}
Olga Russakovsky, Jia Deng, Hao Su, Jonathan Krause, Sanjeev Satheesh, Sean Ma,
  Zhiheng Huang, Andrej Karpathy, Aditya Khosla, Michael~S. Bernstein,
  Alexander~C. Berg, and Fei{-}Fei Li.
\newblock Imagenet large scale visual recognition challenge.
\newblock {\em Int. J. Comput. Vis.}, 115(3):211--252, 2015.

\bibitem{DBLP:conf/iclr/RusuRSVPOH19}
Andrei~A. Rusu, Dushyant Rao, Jakub Sygnowski, Oriol Vinyals, Razvan Pascanu,
  Simon Osindero, and Raia Hadsell.
\newblock Meta-learning with latent embedding optimization.
\newblock In {\em {ICLR} (Poster)}. OpenReview.net, 2019.

\bibitem{DBLP:conf/nips/SchwartzKSHMKFG18}
Eli Schwartz, Leonid Karlinsky, Joseph Shtok, Sivan Harary, Mattias Marder,
  Abhishek Kumar, Rog{\'{e}}rio~Schmidt Feris, Raja Giryes, and Alexander~M.
  Bronstein.
\newblock Delta-encoder: an effective sample synthesis method for few-shot
  object recognition.
\newblock In {\em NeurIPS}, pages 2850--2860, 2018.

\bibitem{DBLP:journals/corr/SimonyanZ14a}
Karen Simonyan and Andrew Zisserman.
\newblock Very deep convolutional networks for large-scale image recognition.
\newblock In {\em {ICLR}}, 2015.

\bibitem{DBLP:conf/nips/SnellSZ17}
Jake Snell, Kevin Swersky, and Richard~S. Zemel.
\newblock Prototypical networks for few-shot learning.
\newblock In {\em {NIPS}}, pages 4077--4087, 2017.

\bibitem{DBLP:conf/cvpr/SungYZXTH18}
Flood Sung, Yongxin Yang, Li Zhang, Tao Xiang, Philip H.~S. Torr, and
  Timothy~M. Hospedales.
\newblock Learning to compare: Relation network for few-shot learning.
\newblock In {\em {CVPR}}, pages 1199--1208. Computer Vision Foundation /
  {IEEE} Computer Society, 2018.

\bibitem{DBLP:conf/nips/VinyalsBLKW16}
Oriol Vinyals, Charles Blundell, Tim Lillicrap, Koray Kavukcuoglu, and Daan
  Wierstra.
\newblock Matching networks for one shot learning.
\newblock In {\em {NIPS}}, pages 3630--3638, 2016.

\bibitem{DBLP:conf/cvpr/WangGHH18}
Yu{-}Xiong Wang, Ross~B. Girshick, Martial Hebert, and Bharath Hariharan.
\newblock Low-shot learning from imaginary data.
\newblock In {\em {CVPR}}, pages 7278--7286. Computer Vision Foundation /
  {IEEE} Computer Society, 2018.

\bibitem{WelinderEtal2010}
P. Welinder, S. Branson, T. Mita, C. Wah, F. Schroff, S. Belongie, and P.
  Perona.
\newblock {Caltech-UCSD Birds 200}.
\newblock Technical Report CNS-TR-2010-001, California Institute of Technology,
  2010.

\bibitem{DBLP:conf/cvpr/XianLSA18}
Yongqin Xian, Tobias Lorenz, Bernt Schiele, and Zeynep Akata.
\newblock Feature generating networks for zero-shot learning.
\newblock In {\em {CVPR}}, pages 5542--5551. Computer Vision Foundation /
  {IEEE} Computer Society, 2018.

\bibitem{DBLP:conf/cvpr/XianSSA19}
Yongqin Xian, Saurabh Sharma, Bernt Schiele, and Zeynep Akata.
\newblock {F-VAEGAN-D2:} {A} feature generating framework for any-shot
  learning.
\newblock In {\em {CVPR}}, pages 10275--10284. Computer Vision Foundation /
  {IEEE}, 2019.

\bibitem{DBLP:conf/aaai/XueW20}
Wanqi Xue and Wei Wang.
\newblock One-shot image classification by learning to restore prototypes.
\newblock In {\em {AAAI}}, pages 6558--6565. {AAAI} Press, 2020.

\bibitem{DBLP:conf/iclr/YangLX21}
Shuo Yang, Lu Liu, and Min Xu.
\newblock Free lunch for few-shot learning: Distribution calibration.
\newblock In {\em {ICLR}}. OpenReview.net, 2021.

\bibitem{DBLP:conf/cvpr/ZhangLYHZ21}
Baoquan Zhang, Xutao Li, Yunming Ye, Zhichao Huang, and Lisai Zhang.
\newblock Prototype completion with primitive knowledge for few-shot learning.
\newblock In {\em {CVPR}}, pages 3754--3762. Computer Vision Foundation {IEEE},
  2021.

\bibitem{DBLP:conf/cvpr/ZhangCLS20}
Chi Zhang, Yujun Cai, Guosheng Lin, and Chunhua Shen.
\newblock Deepemd: Few-shot image classification with differentiable earth
  mover's distance and structured classifiers.
\newblock In {\em {CVPR}}, pages 12200--12210. Computer Vision Foundation /
  {IEEE}, 2020.

\bibitem{DBLP:conf/cvpr/ZhangKJLT21}
Hongguang Zhang, Piotr Koniusz, Songlei Jian, Hongdong Li, and Philip H.~S.
  Torr.
\newblock Rethinking class relations: Absolute-relative supervised and
  unsupervised few-shot learning.
\newblock In {\em {CVPR}}, pages 9432--9441. Computer Vision Foundation /
  {IEEE}, 2021.

\bibitem{DBLP:conf/nips/ZhangCGBS18}
Ruixiang Zhang, Tong Che, Zoubin Ghahramani, Yoshua Bengio, and Yangqiu Song.
\newblock Metagan: An adversarial approach to few-shot learning.
\newblock In {\em NeurIPS}, pages 2371--2380, 2018.

\bibitem{DBLP:conf/icpr/ZhuangTYMZJX18}
Yueqing Zhuang, Li Tao, Fan Yang, Cong Ma, Ziwei Zhang, Huizhu Jia, and
  Xiaodong Xie.
\newblock Relationnet: Learning deep-aligned representation for semantic image
  segmentation.
\newblock In {\em {ICPR}}, pages 1506--1511. {IEEE} Computer Society, 2018.

\end{thebibliography}
}


\end{document}